\documentclass{article}
\usepackage{spconf,amsmath,epsfig}
\usepackage{hyperref}

\usepackage{cite} 
\usepackage{color}

\title{Convolutional Neural Networks for Multispectral Image \\Cloud Masking}

\name{Gonzalo Mateo-Garc{\'i}a, Luis G{\'o}mez-Chova, Gustau Camps-Valls 
\thanks{This work has been partially supported by the Spanish Ministry of Economy and Competitiveness (MINECO, TEC2016-77741-R, ERDF), 
the European Space Agency (ESA IDEAS+ research grant, CCN008), 
and ERC Consolidator Grant SEDAL ERC-2014-CoG 647423. 
\newline
Preprint corresponding to the paper published in 2017 IEEE International Geoscience and Remote Sensing Symposium (IGARSS), Fort Worth, TX, USA, pp. 2255-2258, DOI: 10.1109/IGARSS.2017.8127438.
}}
\address{Image Processing Laboratory (IPL), University of Valencia, Spain}

\begin{document}

\maketitle

\begin{abstract}
Convolutional neural networks (CNN) have proven to be {\it state of the art} methods for many image classification tasks and their use is rapidly increasing in remote sensing problems. One of their major strengths is that, when enough data is available, CNN perform an {\it end-to-end} learning without the need of custom feature extraction methods. In this work, we study the use of different CNN architectures for cloud masking of Proba-V multispectral images. We compare such methods with the more classical machine learning approach based on feature extraction plus supervised classification. Experimental results suggest that CNN are a promising alternative for solving cloud masking problems.
\end{abstract}


\section{Introduction}\label{sec:intro}

In the last years, convolutional neural networks (CNN) have become one of the most promising methods for both general image classification tasks~\cite{NIPS2012_4824,He16} and also remote sensing image classification~\cite{Lngkvist2016ClassificationAS,CastelluccioPSV15,rs71114680}. 
Beyond the high classification accuracy shown in many problems, CNN present interesting properties for remote sensing image processing since they directly learn from the available data the most relevant spatial features for the given problem, i.e. a previous custom feature extraction step is not required~\cite{Romero16}.
In this paper, we analyze the applicability of different CNN architectures in a complex remote sensing problem in which the spatial context is of paramount importance: cloud masking of  Proba-V multispectral imagery.

Images acquired by the Proba-V instrument~\cite{Dierckx14}, which works in the visible and infrared (VIS-IR) ranges of the electromagnetic spectrum, may be affected by the presence of clouds.
Cloud masking can be tackled as a two-class classification problem; and the simplest approach to cloud detection in a scene is the use of a set of static thresholds (e.g. over reflectance or temperature) applied to every pixel in the image, which provides a cloud flag (binary classification). 
However, Proba-V instrument presents a limited number of spectral bands (Blue, Red, NIR and SWIR) which makes cloud detection particularly challenging since it does not present thermal channels or a dedicated cirrus band.
Current Proba-V cloud detection uses multiple thresholds applied to the blue and the SWIR spectral bands~\cite{Lisens00}, but the definition of global thresholds is practically impossible. Hence, for next Proba-V reprocessing~\cite{Wolters15}, monthly composites of cloud-free reflectance in the blue band are used to define dynamic thresholds depending on the land cover type.
In this context, given the reduced amount of spectral information, spatial information seems crucial to increase the performance of classification methods and the cloud detection accuracy.

\section{Methodology}\label{sec:method}

CNN are a special type of neural networks that present a series of convolutional layers especially designed to cope with inputs in the form of multidimensional arrays (image patches)~\cite{LeCun2015}. The CNN architecture used in this work is based on \cite{ChatfieldSVZ14} and consists of 2 blocks of 2 convolutional layers followed by a max-pooling layer. Each convolutional layer is formed by convolution, batch normalization~\cite{BatchNormIoffeS15}, and a rectified linear unit (ReLU). At the top a fully connected (FC) block with 256 hidden units is included, whose outputs are used to predict the output with a sigmoid activation function.

The network is trained to minimize the binary cross-entropy between predictions $h({\bf x}_i,{\boldsymbol \omega})$ and corresponding labels $y_i$:
\[
-\sum_{i=1}^N \Big(y_i \log(h({\bf x}_i,{\boldsymbol \omega})) + (1-y_i) \log(1-h({\bf x}_i,{\boldsymbol \omega})) \Big) ,
\]
where $N$ is the number of training samples, ${\bf x}_i$ is the $i$th input training sample (image patch), $h(\cdot)$ is the network output, ${\boldsymbol \omega}$ is the set of weights of the network, and $y_i$ is the desired output for the image patch central pixel, which will be 1 for cloud contaminated samples and 0 otherwise.
In case of a patch output, i.e. when an entire patch is predicted at a time, the objective is the mean of the binary cross-entropy over all outputs. 
The network was trained with the Adam algorithm~\cite{AdamKingmaB14}, which is a mini-batch stochastic gradient descent algorithm with adaptive estimates of lower order moments.

Finally, a common problem of classification algorithms, and of CNN in particular, is the overfitting problem that produces a poor generalization. The close relationship between the complexity of the classifier and the size of the training set suggests the idea of imposing some kind of regularization when training the models.
To avoid overfitting the {\em dropout technique}~\cite{Dropoutsrivastava14a} is applied at the end of both max-pooling stages (0.25 probability) and after the FC layer (0.5 probability).
In addition, {\em data augmentation} is also employed to increase the size of the training set
by adding flipped versions (left to right and up to down) of the available training samples (image patches). The data augmentation approach is shown in Fig.~\ref{fig:patch_to_patch}, which was previously used in the context of SVMs~\cite{IzquierdoGRSL13}.

\section{Experimental Setup}\label{sec:setup}

\begin{figure}[t]
\begin{center}
\begin{tabular}{cc}
  False RGB &  Output patch labels  \vspace{-0.0cm}\\
 \includegraphics[width=3.8cm]{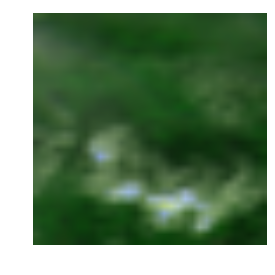} & 
 \includegraphics[width=3.8cm]{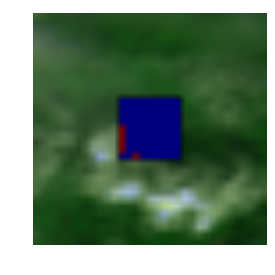} \\
 \includegraphics[width=3.8cm]{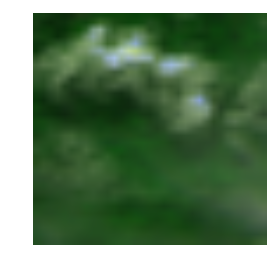} &
\includegraphics[width=3.8cm]{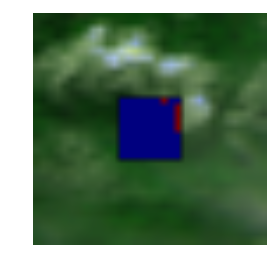}
\\
\end{tabular}
\end{center}
\caption{ \small Example of patch-to-patch input and output with and without data augmentation. In the first row we have the 33x33 patch and the patch with the 9x9 ground truth. In the second row the same patch is flipped and with its ground truth equally flipped \label{fig:patch_to_patch}}
\end{figure}

Two different approaches are analyzed to apply the proposed CNN to the cloud detection of Proba-V:
\begin{itemize}
\item {\it Patch-to-pixel} classification scheme~\cite{Kampffmeyer_2016}: Small patches (subimages with the 4 Proba-V channels) are extracted from the image and used as input data, being the whole patch labeled according to the label of the center pixel. We test two input configurations: 4-channel $17\times17$ and $33\times33$ patches.
\item {\it Patch-to-patch} classification scheme: Instead of classifying the center pixel we classify a central patch. Again we try 4-channel $17\times17$ and $33\times33$ patches as inputs and predict a $9\times9$ output patch. Figure \ref{fig:patch_to_patch} shows input and output patches as they are fed to the network.
\end{itemize}
Bigger input patch sizes make the model slower to train and to run. This is an important issue since cloud masking algorithms should be applied to all images acquired by the satellite. On the other hand, bigger input sizes allow the model to integrate more surrounding information, which could make the model more accurate. With the output patch sizes the trade-off happens the other way around: bigger output sizes will make the model faster since we predict an entire patch at a time instead of a pixel. 

All CNN models were implemented in {\em Python} using the {\em Keras} library~\cite{KerasChollet2015}. Training was done in CPUs for small models (input size $17$) and GPUs for bigger ones ($33\times33$ input). Training time ranged from 20 to 30 hours depending on the load of the computer. 

In the experiments, we benchmark the proposed CNN approaches against two {\it state of the art} classifiers: standard fully connected neural networks (multilayer perceptron, MLP) and gradient boosting machines (GBM) \cite{friedman2001,XGboostChenG16}.
We trained both methods following a {\it pixel-to-pixel} classification scheme with different information content at the used input data: (1) the four channels of the instrument ({\it bands}); (2) the four channels plus ten spectral features useful for cloud detection ({\it feat}); and (3) the channels plus spectral features plus basic spatial features ($3\times3$ and $5\times5$ mean and std) from which we finally select 40 relevant spatio-spectral inputs (denoted by {\it all}).

\section{Experimental Results}\label{sec:results}

The available data set consists of 60 Proba-V images acquired in four days covering the four seasons: 21/03/2014, 21/06/2014, 21/09/2014, and 21/12/2014. All models are trained on 100,000 pixels randomly chosen from these images. An independent set of 360,000 pixels was left for testing purposes. The datasets were balanced to contain an equal number of cloud-contaminated and cloud-free pixels. 

First, we look at the reference GBM and MLP machine learning approaches (Table~\ref{tab:ml_acc}). The effect of feature extraction results in a boost in the performance, specially when spatial information is included. Additionally, we notice that both GBM and MLP achieve similar accuracies on the test set. 

\begin{table}[t]
\caption{ \small Gradient boosting machines and neural networks accuracies on {\it pixel to pixel} classification scheme using as inputs: (1) the bands ({\it bands}); (2) bands and spectral features ({\it feat}); and (3) bands, spectral features and spatial features ({\it all}). \label{tab:ml_acc}}
\begin{center}
\begin{tabular}{c|cc}
\hline\hline
{\bf Inputs} & {\bf GBM} & {\bf MLP} \\
\hline
(1) {\it bands} & 92.92\% & 93.43\% \\
(2) {\it feat} &  93.39\% & 93.51\% \\
(3) {\it all} & {\bf 94.60}\% & {\bf 94.51}\% \\
\hline\hline
\end{tabular}
\end{center}
\end{table}

\begin{table}[t]
\caption{ \small Train and test set accuracy with and without data augmentation  on the $33\times33$ {\it patch to pixel}  CNN model. The model with data augmentation incurs in less overfitting. \label{tab:data_aug}}
\begin{center}
\small
\begin{tabular}{c|cc}
\hline\hline
{\bf Data Augmentation}  & {\bf Train} (OA / Kappa) & {\bf Test} (OA / Kappa) \\
\hline
{\bf NO}  & 98.55\% / 0.9709 &  95.05\% / 0.9007 \\
{\bf YES} & 96.11\% / 0.9221 & {\bf 95.44}\% / {\bf 0.9085} \\
\hline\hline
\end{tabular}
\end{center}
\end{table}

\begin{figure*}[t]
\begin{center}
\setlength\tabcolsep{-6.5pt}
\begin{tabular}{cc}
$17\times17$ input patch to $9\times9$ output patch &  $33\times33$  input patch to $9\times9$ output patch  \vspace{-0.0cm}\\
\includegraphics[width=9cm]{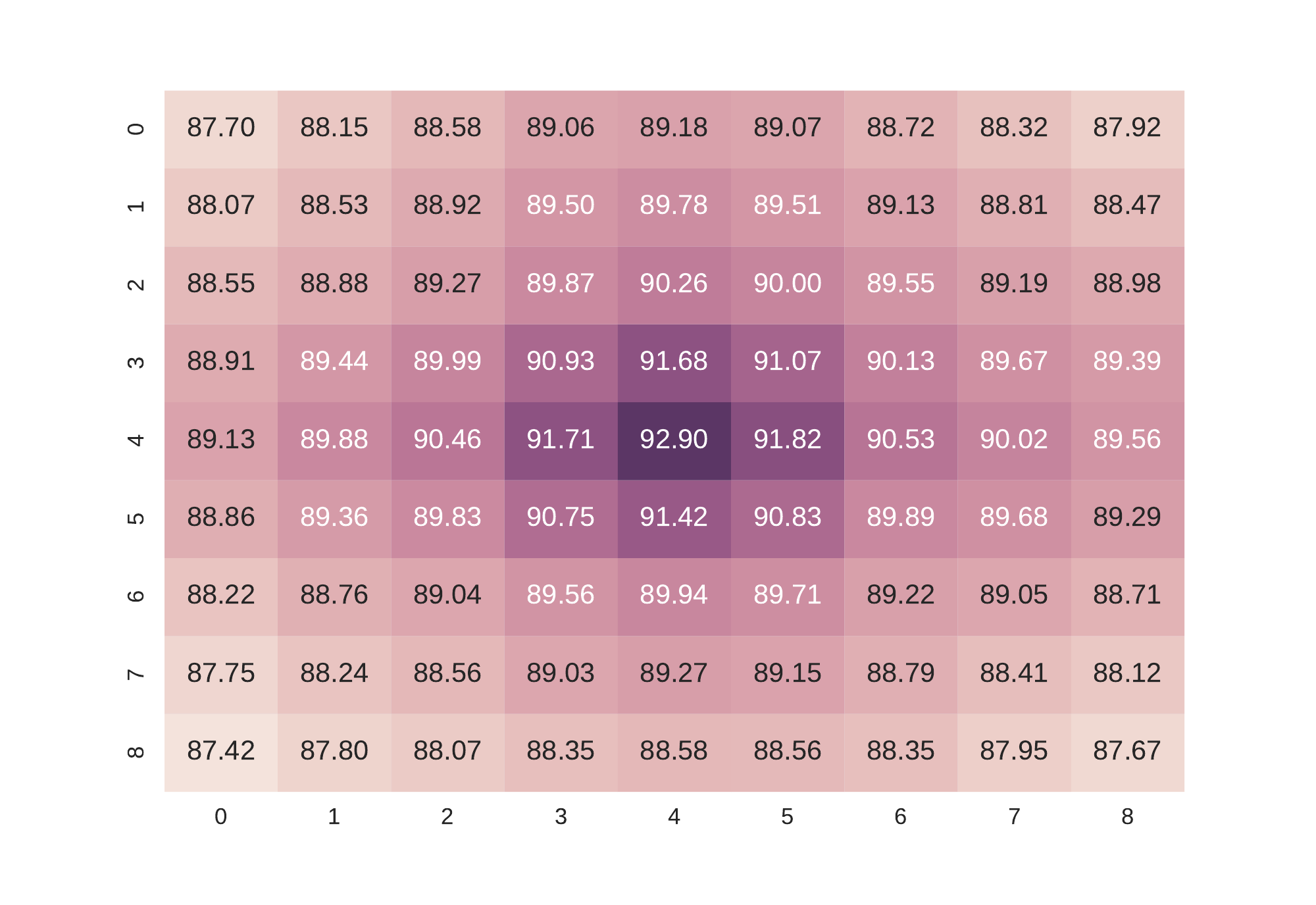} & 
\hspace*{-0.5cm}
\includegraphics[width=9cm]{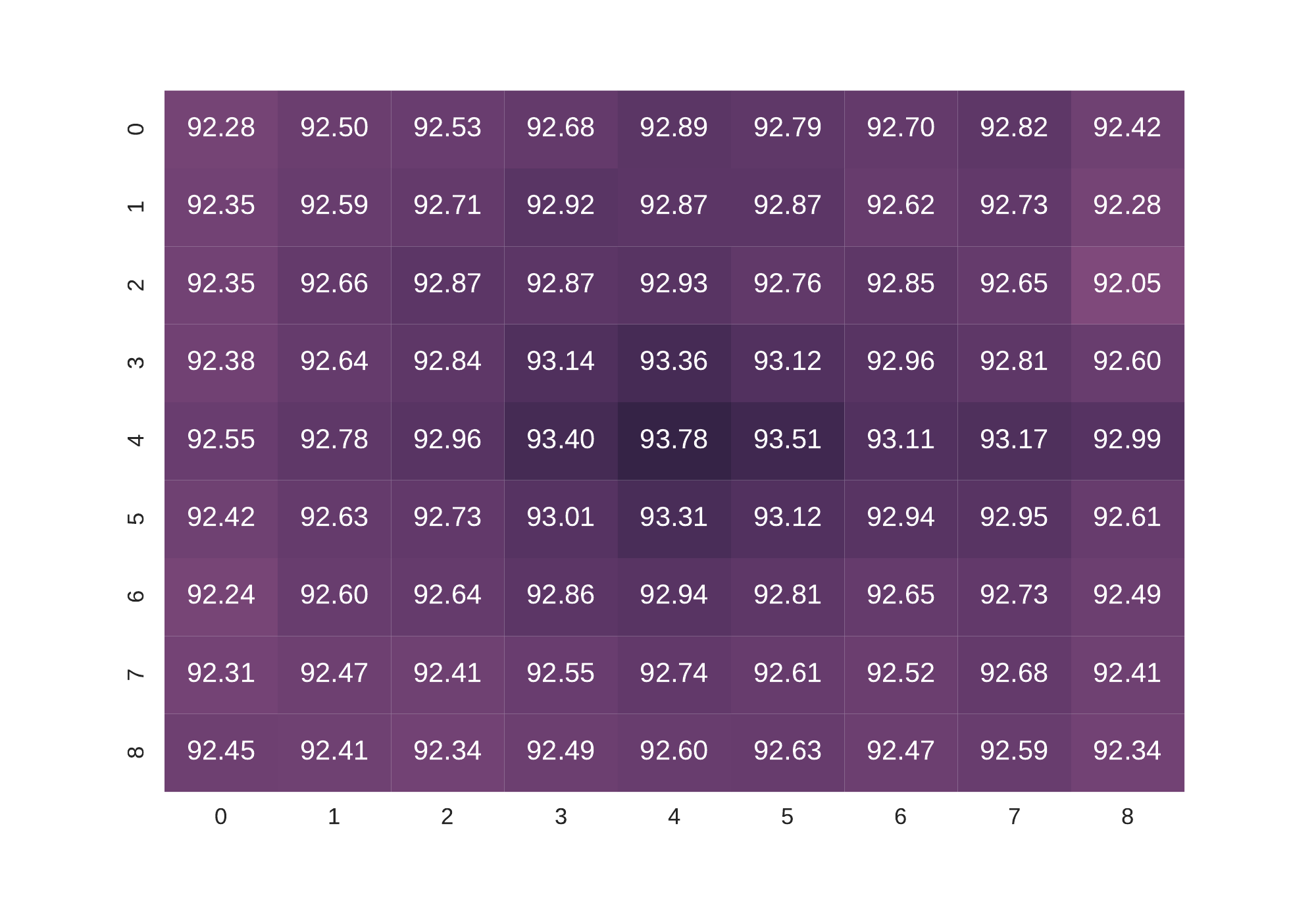} \\
\end{tabular}
\end{center}
 \vspace*{-0.8cm}
\caption{ \small Detection accuracy (\%) of {\it patch-to-patch} models per pixel of the $9\times9$ output patch. \label{fig:acc_patch_to_patch}}
\end{figure*}

In the case of CNN, first we confirm in Table~\ref{tab:data_aug} the improvement, due to the proposed data augmentation, on the the overall accuracy (OA\%) and Cohen's Kappa statistic ($\kappa$).
Then, Table~\ref{tab:cnn_acc} shows that both CNN ($17\times17$ and $33\times33$ input) outperform the classical approach of {\em feature extraction} plus {\em supervised classification} when we predict the central pixel ({\it patch to pixel}), while {\it patch to patch} approaches result in lower accuracies.
Figure~\ref{fig:acc_patch_to_patch} shows the accuracy over the whole $9\times9$ output patch, where one can observe that central pixels are more accurately predicted whereas accuracies over pixels on the boundaries of the $9\times9$ patches are lower.

\begin{table}[t]
\caption{ \small CNN accuracies of different input/output configurations. Patch to patch accuracies are measured in the center pixel and the mean over all patches overlapping this pixel. \label{tab:cnn_acc}}
\begin{center}
\begin{tabular}{c|c|cc}
\hline\hline
 & {\it patch to pixel} & \multicolumn{2}{c}{{\it patch to patch}} \\
 & &  center & mean \\
\hline
$17\times17$ input & 94.92\% & 92.90\% & 90.33\% \\
$33\times33$ input & {\bf 95.44}\% & 93.78\% & 93.06\% \\
\hline\hline
\end{tabular}
\end{center}
\end{table}

\begin{figure}[t]
\begin{center}
 \includegraphics[width=8.5cm]{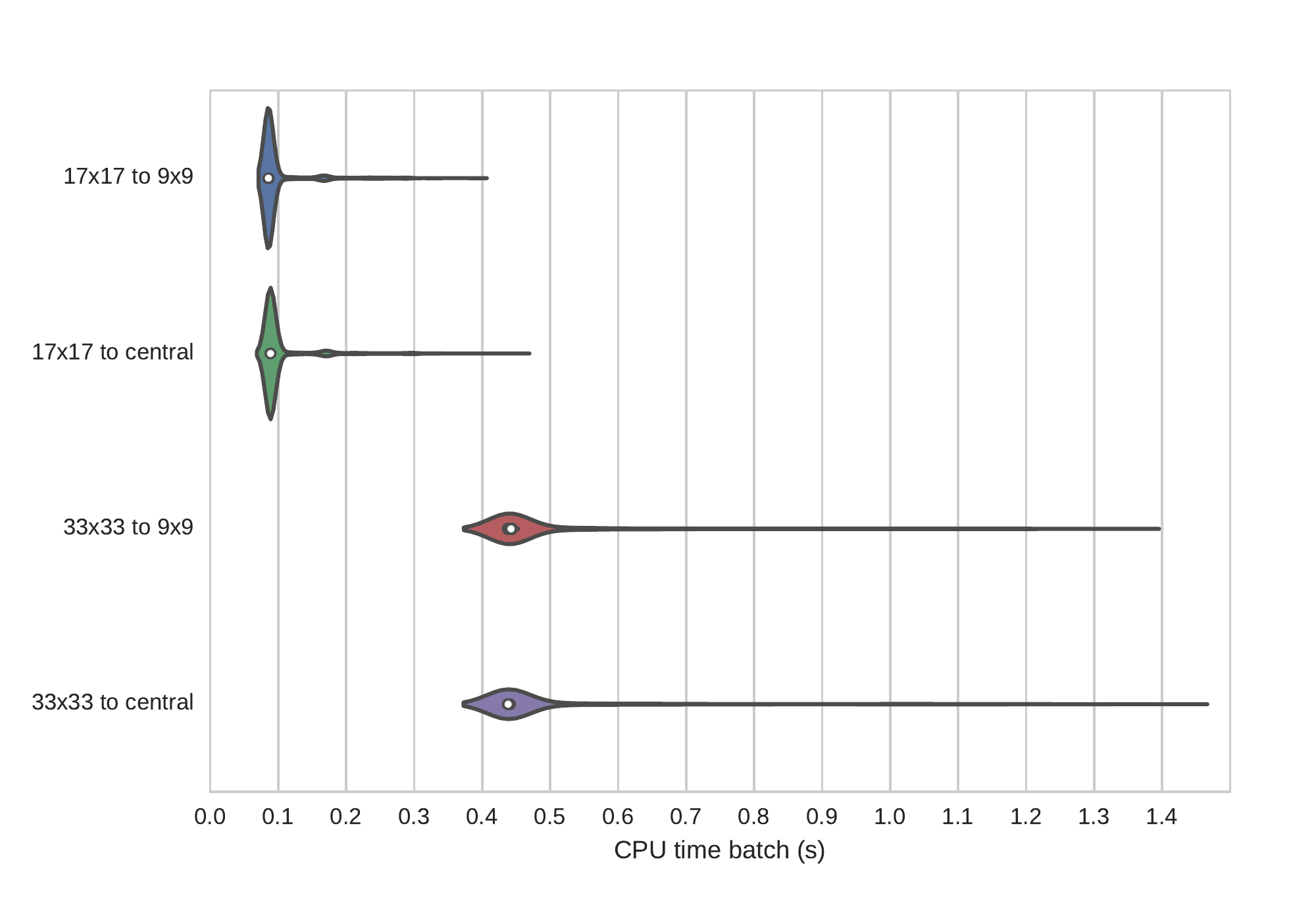}
 \end{center}
\vspace{-0.5cm}
\caption{ \small CNN computational cost per batch of 128 patches using CPUs. Times measured over 2970 different batches. \label{fig:cnn_time}}
\end{figure}

It should be noticed that the networks trained in this work are smaller, in terms of the number of weights, than common CNN presented in the literature~\cite{Kampffmeyer_2016}. However, our problem can be considered less complex since there are only two classes. In addition, having a smaller network presents the advantage of a lower computational cost during the test phase.
Figure \ref{fig:cnn_time} shows the computational time of the proposed CNN models per batch of 128 4-channel image patches. As we discussed before, smaller input sizes result in lower computational cost of the whole network. In addition, the burden of predicting a patch instead of a single value is barely noticed.
Since {\it patch-to-patch} prediction yields a $9\times9$ output, patch prediction is 81 times faster than predicting the center pixel for a complete image. Nevertheless accuracy is reduced from 95.44\% to 93.06\% in the case of $33\times33$ input size (see Table~\ref{tab:cnn_acc}). When choosing a $17\times17$ input size, accuracy loss is higher, dropping from 94.92\% to 90.33\%.

\begin{figure}[h]
\begin{center}
\footnotesize
\begin{tabular}{cc}
RGB Composite (2014/06/21) & Predicted Cloud Mask \\
\hspace{-0.5cm}
\includegraphics[width=4.5cm]{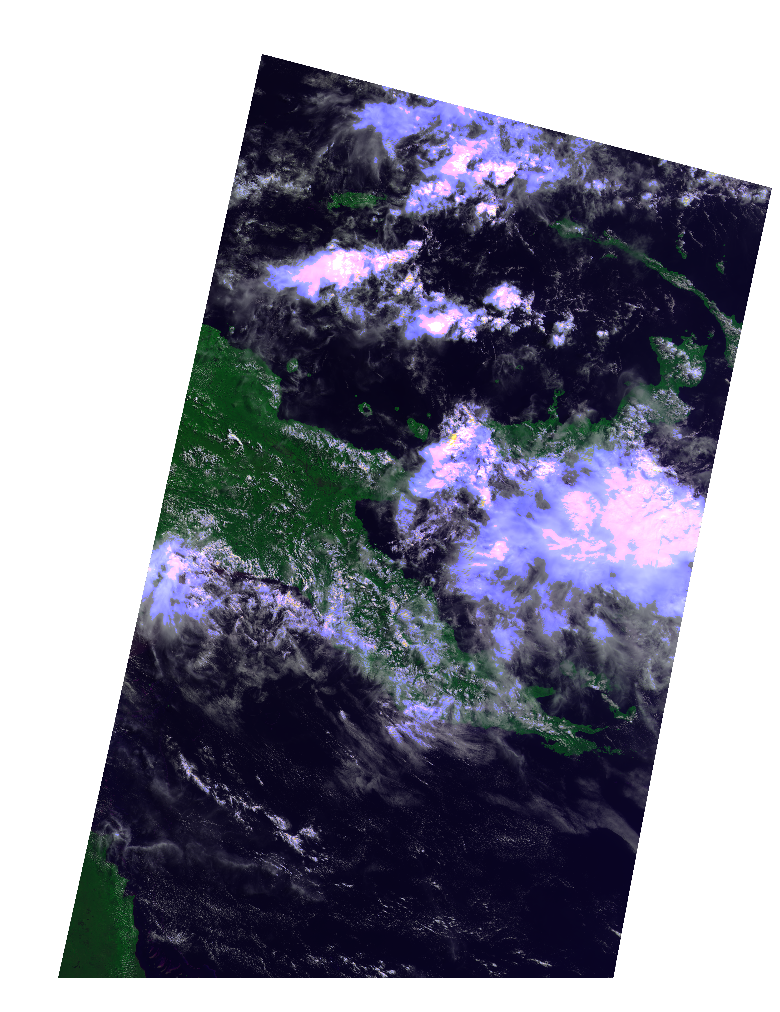} &
\hspace{-1.0cm}
\includegraphics[width=4.5cm]{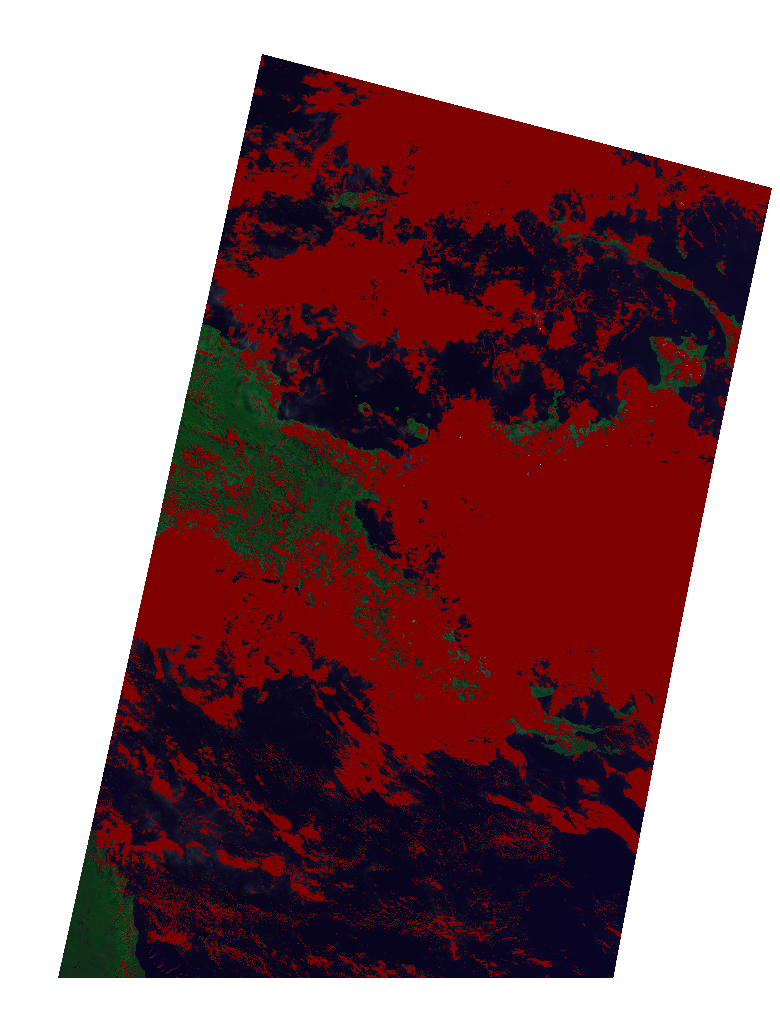}
\end{tabular}
\end{center}
\caption{ \small \label{fig:cloudmask_example}   Example showing the RGB false color composite and the cloud mask obtained with the $33\times33$ {\it patch to pixel} CNN.}
\end{figure}

Finally, an illustrative example of the resulting cloud mask is shown in Fig.~\ref{fig:cloudmask_example}. In this figure, we show an scene of Papua New Guinea with a complex cloud structure over ocean and land.
The high overall detection accuracy (95\%) and Cohen's Kappa statistic ($\kappa$=0.87) confirm the visual agreement between the cloud pattern and the obtained cloud mask.

\section{Conclussions}\label{sec:conclussions}

In this paper, we presented a comprehensive study of the application of CNN models to cloud masking of Proba-V satellite images. We shown that CNN models outperform the classical approach of {\em feature extraction} plus {\em supervised classification}, even using advanced machine learning methods, in this cloud detection problem. We compared different input and output network configurations (patch sizes) that revealed a trade-off between classification accuracy and computational cost.

Future work is tied to better analyze the CNN training hyperparameter selection and to study the detection performance over high reflectance surfaces such as ice/snow, sand and urban areas. Another direction is to couple cloud and shadow detection on the CNN classifier.

\small
\bibliographystyle{IEEEbib}
\bibliography{}

\end{document}